# The role of inhibitory control in garden-path sentence processing: A Chinese-English bilingual perspective

Xiaohui Rao[1,2], Haoze Li[3], Xiaofang Lin[1], Lijuan Liang*[1]

*[1] Bilingual Cognition and Development Lab, Centre for Linguistics and Applied Linguistics, Guangdong University of Foreign Studies; [2] Department of Linguistics and Modern Languages, The Chinese University of Hong Kong; [3] Division of Linguistics and Multilingual Studies, Nanyang Technological University*

**Acknowledgement:**

The authors gratefully acknowledge the invaluable advice provided by Prof. Ziyin Mai, as well as the assistance of Muxin Lin and Luyao Wang during data collection.

**Address for correspondence:**

Lijuan Liang
Bilingual Cognition and Development Lab
Center for Linguistics and Applied Linguistics
Guangdong University of Foreign Studies
Guangzhou, 510420, China. Telephone: 86-18124239738
E-mail: lianglj@gdufs.edu.cn

**Abstract**

In reading garden-path sentences, people must resolve competing interpretations, though initial misinterpretations can linger despite reanalysis. This study examines the role of inhibitory control (IC) in managing these misinterpretations among Chinese-English bilinguals. Using self-paced reading tasks, we investigated how IC influences recovery from garden-path sentences in Chinese (L1) and its interaction with language proficiency during English (L2) processing. Results indicate that IC does not affect garden-path recovery in Chinese, suggesting reliance on semantic context may reduce the need for IC. In contrast, findings for English L2 learners reveal a complex relationship between language proficiency and IC: Participants with low L2 proficiency but high IC showed lingering misinterpretations, while those with high proficiency exhibited none. These results support and extend the Model of Cognitive Control (Ness et al., 2023). Moreover, our comparison of three Stroop task versions identifies L1 colour-word Stroop task as the preferred measure of IC in bilingual research.

## 1. Introduction

Sentence processing does not always proceed incrementally (Demberg & Keller, 2019), sometimes necessitating reanalysis and even resulting in lingering misinterpretations (Christianson et al., 2001). A typical example is the processing of garden-path sentences. Consider (1).

(1) *While the man hunted the deer ran through the woods.*

In this sentence, the initial interpretation is that *the deer* is the object of *hunted*. However, upon reaching the subsequent verb *ran*, the comprehender must revise this initial misinterpretation and reanalyze the structure, recognizing that *the deer* is actually the subject of *ran* (Christianson et al., 2001, 2006; Ferreira & Henderson, 1991; Frazier & Rayner, 1982; Fujita & Cunnings, 2020; Malyutina & den Ouden, 2016; Van Gompel et al., 2006). The extra cognitive effort exerted during reanalysis is called the garden-path effect (Ferreira & Henderson, 1990; Frazier & Rayner, 1982). Furthermore, even after reanalysis is completed, misinterpretations can persist (Christianson et al., 2001, 2006; Fujita & Cunnings, 2020; Jacob & Felser, 2016; Malyutina & den Ouden, 2016; Nakamura & Arai, 2016; Slattery et al., 2013; Sturt, 2007; Van Gompel et al., 2006). Notably, garden-path recovery involves a competition between two syntactic structures, requiring the inhibition of the incorrect interpretation to arrive at the correct one. Previous research has demonstrated substantial evidence for the role of inhibitory control in native language (L1) processing, supported by both behavioural studies (Hsu & Novick, 2016; May & Scofield, 2024; Trueswell et al., 1999; Vuong & Martin, 2014) and neuroimaging studies (Fedorenko, 2014; January et al., 2009); however, most studies have focused on Indo-European languages, particularly English (Choi & Trueswell, 2010). In contrast, there is a lack of research on the role of inhibitory control in Chinese garden-path sentences, which possess unique syntactic features. A deeper investigation into this area could elucidate inhibitory control mechanisms underlying temporary syntactic ambiguity resolution cross-linguistically. Moreover, while inhibitory control's role in native English (L1) garden-path recovery is well-established, its function during second language (L2) processing, especially its interaction with L2 proficiency, remains unclear and warrants further examination. Therefore, this study aims to address these two critical gaps concurrently by examining the processing of garden-path sentences among Chinese-English bilinguals.

## 1.1 Lingering misinterpretations and competing mechanisms

Although the comprehender can engage in reanalysis and ultimately interpret the sentence correctly, Christianson et al.'s (2001) pioneer study demonstrates that the original incorrect analysis lingers even after reanalysis takes place. In their experiments, the participants were asked to read the garden-path sentences like (1) and then answer forced-choice (yes/no) questions like those below.

(2a) *Did the man hunt the deer?*

(2b) *Did the deer run through the woods?*

The results were intriguing. While the participants answer "yes" (correctly) to (2b) nearly 90% of the time, they also answer "yes" (incorrectly) to (2a) around 60% of the time. Given that the sum of the proportions exceeded one, Christianson et al. concluded that the participants still believed incorrectly that the man hunted the deer, though they understood that the deer ran through the woods. In this sense, the misinterpretation persisted to some extent after the processing ended. To explain the persistence of misinterpretation, the Good Enough Language Processing model has been proposed and continuously updated (Christianson et al., 2001; Ferreira et al., 2001, 2002; Ferreira & Patson, 2007; Huang & Ferreira, 2021; Qian et al., 2018; Slattery et al., 2013; Sturt, 2007), developing into the following two alternative accounts (3a, 3b).

(3a) *Incomplete reanalysis account:* Parsers fail to establish a fully specified correct structure after reanalysis, and only rely on semantic-based, or good-enough, representations to interpret garden-path sentences.

(3b) *Unerased false memory trace account:* Parsers complete syntactic reanalysis by constructing correct structures for garden-path sentences, but they fail to fully erase the memory traces of initial misinterpretations.

There has been increasing evidence supporting the *unerased false memory trace account* (Cunnings, 2017; Fujita & Cunnings, 2020; Slattery et al., 2013; Sturt, 2007; Qian et al., 2018). More importantly for the current research, both reanalysis and lingering misinterpretations may be closely tied to competing mechanisms. Specifically, reanalysis involves a competition of two syntactic structures, whereas the lingering misinterpretation results from a competition of two interpretations. When processing garden-path sentences, readers need to suppress the activation of initial structural analyses and local interpretations in favour of globally correct alternatives. It is worth noting that the output of such competing processes doesn't seem constant. In particular, lingering misinterpretations vary across trials. It is possible that in some trials, readers can better inhibit the initial misinterpretation, while in other trials, the competition between interpretations results in the persistence of the initial misunderstanding. This raises an important question: What factors influence these competing processes?

Such competitions may rely on a specific cognitive ability: inhibitory control. Inhibitory control is a core cognitive control managing human beings' automatic urges by pausing, then using attention and reasoning to respond appropriately (Aron et al., 2014; Munakata et al., 2011). Specifically, it is essential for halting or redirecting cognitive processes; without it, the flexibility to adapt mental strategies and behaviours would be compromised (Anderson, 2005). Inhibitory control is applicable in various contexts, such as overriding habitual responses and selectively targeting memories to regulate memory retrieval. It can be assessed through tasks like the Stroop task, where participants suppress the impulse to read a conflicting written word (e.g., the word blue printed in the red colour) and respond with the colour (e.g., Jensen & Rohwer, 1966; Scarpina & Tagini, 2017). This task requires participants to inhibit the literal meaning and focus on the real colour. Regarding garden-path sentences, inhibitory control may subserve both reanalysis and the management of unwanted memory traces. Specifically, it likely functions to reduce the activation of an initial misinterpretation, while concurrently enhancing an alternative, contextually-appropriate analysis (Ness et al., 2023).

### 1.2 The need to investigate inhibitory control in Mandarin Chinese (L1)

Empirical evidence supports a relationship between inhibitory control and garden-path sentence processing among L1 speakers, predominantly English (Hsu & Novick, 2016; May & Scofield, 2024; Trueswell et al., 1999; Vuong & Martin, 2014). This connection is evident in children. Studies employing the visual world paradigm have shown that compared to adults, children were slower and less likely to look at the correct referent, indicating that they struggle more with recovering from garden-path sentences, often failing to revise their initial interpretations and holding onto incorrect parses (Trueswell et al., 1999). This developmental asymmetry is likely attributed to children's comparatively underdeveloped inhibitory control abilities, as children with stronger inhibitory control show fewer difficulties in processing garden-path sentences (May & Scofield, 2024). Nevertheless, other age-related factors, such as language proficiency, cannot be excluded as potential contributors to this phenomenon.

Besides, there is some evidence for the role of inhibitory in native adults.

Inhibitory control in this research area has been assessed using various tasks and paradigms (see Methods for more details). For instance, certain studies utilising a self-paced reading paradigm have found that individuals with higher inhibitory control, as measured by the verbal Stroop task, were more successful in recovering from misinterpretations, reflected by reduced response time at the sentence-final region and improved accuracy in the grammaticality judgements (Vuong & Martin, 2014). Besides, improvements in reanalysis and reduced lingering misinterpretations have been observed immediately after participants completed inhibitory control tasks in a cross-task-adaptation paradigm (Hsu & Novick, 2016), which indicates that sustained engagement of inhibitory control facilitates the processing of garden-path sentences. Collectively, prior empirical findings suggest that inhibitory control facilitates reanalysis and mitigates lingering misinterpretations by inhibiting the activation of initial wrong interpretations while promoting the activation of globally correct interpretations.

It is important to note that the effect size of inhibition may depend on the type of ambiguity present in sentences. Some studies indicate that certain temporary ambiguity structures, such as object/subject ambiguity (e.g., Ferreira et al., 2001), are more difficult to process. In contrast, coordination ambiguities, exemplified by the sentence *Put the butter in the bowl and the pan on the towel*, are less complex (Bailey & Ferreira, 2003; Fodor & Inoue, 1998; Staub & Clifton, 2006). This variation in ambiguity structures may result in differing levels of competition and affect how easily the processing system can recover from misinterpretations (Engelhardt & Ferreira, 2010). Moreover, Sturt (2007) found that full reanalysis was more likely conducted when a semantic cue, like plausibility information, was present in the sentence. Taken together, it can be suggested that the strength of ambiguity can substantially affect the competition involved in recovering from temporary ambiguities, which likely influences the functioning of inhibitory control during the comprehension of garden-path sentences. Therefore, the predominance of studies that have examined the processing of garden-path sentences in English may provide a limited perspective on how people process garden-path sentences (Choi & Trueswell, 2010).

For this reason, an important recent development has been the study of other language families that exhibit huge typological differences from English, such as Chinese. Specifically, English has a rigid grammatical structure with strict rules governing word order, subject-verb agreement, and tense; deviations from these rules can lead to confusion. In contrast, Chinese syntax is more flexible, relying on context and particles like *le* to convey temporal notions rather than inflections. Additionally, there is no clear correspondence between word categories and syntactic components in Chinese. This flexibility allows the processing of Chinese sentences to prioritize semantics over strict syntax, enabling meaning to be inferred from context, while English sentence processing emphasizes syntactic clarity (Chen, 1984; Chu, 1998; Ming, 2023; Yang et al., 2010). Therefore, exploring the ambiguities present in Chinese garden-path sentences may provide valuable insights into the role of inhibitory control in sentence processing across different linguistic contexts (Xu & Huang, 2025).

### *1.3 Disentangling the interplay between inhibitory control and language proficiency in the L2*

Overall, previous findings suggest that inhibitory control plays a certain role in garden-path sentences, but its functioning is complex and may be interconnected with other factors. According to the Model of Cognitive Control proposed by Ness et al. (2023), inhibitory control, broadly referred to as cognitive control in their paper, serves as a biasing mechanism that enhances the activation of interpretations supported by the most reliable evidence while suppressing alternative interpretations to resolve sentence ambiguity. Importantly, they highlight a close relationship between inhibitory control and linguistic knowledge which has been a key component of sentence-processing for decades (Reichle, 2021). They argue that the role of inhibitory control depends on linguistic knowledge, which helps determine the interpretation most strongly supported by the available evidence. Given that linguistic knowledge is closely linked to language proficiency, a critical connection between inhibitory control and language proficiency can be inferred. While this connection is less apparent in L1 research due to the ceiling effect of native proficiency, it may become more evident in L2 contexts, where variability in language proficiency is greater. Thus, the interplay between inhibitory control and language proficiency in processing garden-path sentences warrants further investigation through an L2 perspective.

The influence of inhibitory control on the processing of L2 garden-path sentences has not been extensively explored. Despite its recognized role in English, it is unclear whether this influence extends to L2 English learners. Two recent studies have investigated this issue, but their findings diverge. Xie et al. (2022) required Chinese-English bilinguals with strictly matched L2 proficiency levels to read L2 garden-path sentences in a self-paced manner and do three tasks measuring the subcomponents of cognitive control (i.e., The digit span task measuring working memory, the Wisconsin card sorting test measuring shifting, the Flanker task measuring inhibitory control and conflict monitoring). They concluded that working memory and abilities of conflict monitoring positively correlate with faster reanalysis, while inhibition and shifting groups do not have a significant impact. Additionally, their study observed that cognitive control does not modulate lingering misinterpretations. Conversely, a conference paper by Ma et al. (2020) reported that L2 learners with stronger inhibitory control, measured by nonverbal Stroop task, exhibited faster reading times for garden-path sentences, as evidenced by eye-movement data, though no evidence suggests that inhibition influences lingering misinterpretations.

However, these two studies suffer from two limitations. First of all, both overlooked language proficiency as a potential modulating factor. It is possible that high L2 proficiency among participants may have mitigated the influence of inhibitory control, obscuring its role in the processing of garden-path sentences. Second, the divergent findings between the studies may be attributable to the use of different measures of inhibitory control. Specifically, Ma et al. (2020) employed the Stroop task, whereas Xie et al. (2022) utilised the Flanker task. The different ways that prior research has measured inhibitory control make it hard to compare the findings and draw clear conclusions from the existing research. As such, this study would address these limitations by providing more systematic investigations that account for both L2 proficiency and the measurement of inhibitory control (See the Method section for details). To be specific, this study will take into consideration not only inhibitory control but also L2 proficiency to investigate their interactive effect on the recovery from syntactic ambiguity, as elucidated in the Model of Cognitive Control (Ness et al., 2023). The findings will further shed light on the precise role of inhibitory control in L2 garden-path processing.

## *1.4 The present study*

The present study aims to tap into the role of inhibitory control in garden-path sentence processing on bilinguals whose L1 is Mandarin Chinese and L2 is English. Our two objectives were: 1) to investigate the role of inhibitory control in the processing of garden-path sentences in Chinese, a non-inflectional and analytic language; and 2) to explore the relationship between inhibitory control and language proficiency from an L2 perspective. By conducting experiments with a homogeneous group of Chinese-English bilinguals, we aimed to obtain comparable evidence regarding the dynamic role of inhibitory control as it changes with language typology and language proficiency. Specifically, we address two core questions: 1) How does inhibitory control modulate the processing of garden-path sentences in Chinese (L1)? 2) How does inhibitory control interact with language proficiency in modulating the processing of garden-path sentences in English (L2)?

To explore these questions, we employed self-paced reading tasks to examine the processing of garden-path sentences in both Chinese (L1) and English (L2). Based on prior research, we consider the reaction times (RTs) for the reanalysis region as an index of garden-path reanalysis (i.e., garden-path effects) and the accuracies (ACCs) of the probe sentences as an indicator of lingering misinterpretations (e.g. Christianson et al., 2001, 2006; Fujita & Cunnings, 2020; Jacob & Felser, 2016). We analysed the effect of inhibitory control in both RT and ACC analyses. We also applied three Stroop tasks (i.e., L1 colour-word Stroop task, L2 colour-word Stroop task, Number Stroop task) to assess Chinese-English bilinguals' inhibitory control abilities (See the Method section for details) and the Oxford Quick Placement Test to measure their L2 proficiency.

Our primary hypotheses are: 1) Inhibitory control influences the processing of garden-path sentences in Chinese (L1); 2) Inhibitory control influences the processing of garden-path sentences in English (L2), with L2 proficiency potentially acting as a moderating factor.

## 2. Methods

### 2.1 Participants

Forty-two (mean age = 20 years old, range = 18-24; 2 males) college students, who were all native speakers of Mandarin Chinese, participated in this study. All started acquiring English around seven years old ($M = 7.21$, $SD = 2.72$), and have been learning and using English averagely for 12 years ($M = 12.90$, $SD = 2.60$).

All participants reported normal or corrected-to-normal vision, no colour blindness or colour deficiency, right-handedness, and no history of mental illness or speech disorders. They provided informed written consent and received 70 yuan for their participation. During the recruitment session, a screening test was administered to assess participants' abilities to comprehend garden-path sentences in English. This test consisted of single-choice questions requesting participants to choose the best-fitting Chinese translation for one English garden-path sentence, along with three fillers. It is noteworthy that only one garden-path sentence was involved, in order to prevent participants from discerning the experimental intent. Given that all participants successfully passed the test, they demonstrated the capability to achieve globally correct interpretations of garden-path sentences. Note that the purpose of the screening test was not disclosed to the participants.

### 2.2 Materials

*L1 self-paced garden-path sentence reading task*

Since Chinese garden-path sentences do not exhibit object/subject ambiguities like those in English, sixty pairs of sentences in Chinese were created in a manner similar to the study by Xu and Huang (2025), each consisting of sentences in two conditions—garden-path and non-garden-path, as exemplified in Table 1.

**Table 1.** *Sample stimuli for Chinese garden-path sentences.*

| Condition | Target Sentence | | | Probe Sentence |
|---|---|---|---|---|
| | Pre-reanalysis region | Reanalysis region | Post-reanalysis region | |
| garden-path | 小杰(N)/ 喜欢(V)/ 看书(VP)/ <br> Xiaojie/ xihuan/ kanshu/ <br><br> Xiaojie likes read-book <br><br><br> "Xiaojie likes the students who love reading books." | 的/ <br> **de**/ <br><br> MOD | 学生(N)。/ <br> xuesheng/ <br><br> student | a.小杰 喜欢 看书。<br>Xiaojie xihuan kanshu.<br>Xiaojie like read-book<br>"Xiaojie likes reading."<br><br>b.小杰 喜欢 学生 看书。<br>Xiaojie xihuan xuesheng kanshu.<br>Xiaojie like student read-book<br>"Xiaojie likes students reading." |
| non-garden-path | 小杰(N)/ 喜欢(V)/ 风趣(A) / <br> Xiaojie/ xihuan/ fengqu/ <br><br> Xiaojie likes humorous <br><br><br> "Xiaojie likes humorous students." | 的/ <br> **de**/ <br><br> MOD | 学生(N)。/ <br> xuesheng/ <br><br> student | a.小杰 喜欢 保持 风趣。<br>Xiaojie xihuan baochi fengqu.<br>Xiaojie like stay humorous<br>"Xiaojie likes to be humorous."<br><br>b.小杰 喜欢 学生 是 风趣 的。<br>Xiaojie xihuan xuesheng shi fengqu de.<br>Xiaojie like student be humorous SFP.<br>"Xiaojie likes students to be humorous." |

Regarding the stimuli presented on Table 1, the particle *de* is a modifier marker, indicating a modification structure where the preceding phrase *kanshu* "reading" or *fengqu* "humorous" modifies the following noun phrase *xuesheng* "student". However, the garden-path sentence introduces a temporary structural ambiguity. Before the appearance of *de*, *kanshu* "reading" would be parsed as the object of the main verb *xihuan* "like", in line with the fundamental SVO word order of Chinese. When it comes to *de*, parsers would recognize that *kanshu* "reading" should form a modification structure with *xuesheng* "student", rather than functioning as the object, triggering reanalysis. By contrast, the non-garden-path sentence does not evoke a temporary structural ambiguity, as the adjective *fengqu* "humorous" cannot be selected by a verb as the object and hence can only be analysed as the modifier.

All garden-path and control sentences were rated for the degree of plausibility (1-highly implausible; 6-highly plausible) by 20 Chinese-English bilinguals who didn't participate in the formal experimental session. The results showed that all sentences were regarded as plausible (garden-path: $M = 5.37$, $SD = 0.59$; non-garden-path: $M = 5.67$, $SD = 0.26$). In addition, forty filler sentences were also included in the task, including sentences with various types of syntactic structures, to keep participants blind to the purpose of the study.

After reading each target sentence, participants were required to judge whether the probe sentence was congruent with the target sentence. For garden-path sentences, the probe sentences fell into two types, half congruent with the globally correct interpretation and the other half congruent with the initial misinterpretation. The accuracy of probe sentences could reflect participants' reading comprehension and the degree of lingering misinterpretation, a similar design to Christianson et al. (2001). For non-garden-path sentences, half of the probe sentences were consistent with the former part of the experimental sentences and the other half were consistent with the whole sentence. Half of the fillers' probe sentences were congruent with the partial reading of the filler, while the rest were congruent with the global reading of the filler.

Two stimulus lists were formed according to the Latin Square design, with both lists including 30 garden-path sentences, 30 non-garden-path sentences, and 40 fillers. None of the sentences was repeated within each list. Half probe sentences were judged correct and the other half wrong. The assignment of congruent and incongruent responses to the left and right hands was counterbalanced across participants.

*L2 self-paced garden-path sentence reading task*

Sixty pairs (garden-path and non-garden-path) of sentences in English were created, using the typical structure of English garden-path sentences commonly adopted in previous studies (Christianson et al., 2001; Ferreira et al., 2001) (for examples), as shown in Table 2.

**Table 2.** *Example of experimental stimuli in L2 self-paced sentence reading task.*

| Condition | Context | Reanalysis Region | Others | Probe Sentence |
|---|---|---|---|---|
| garden-path sentence | When / the man / <u>hunted</u> / the dog / | **ran** / | into the woods. / | a. The man hunted the dog. b. The dog ran away. |
| non-garden-path sentence | When / the man / <u>read</u> / the dog / | **ran** / | into the woods. / | a. The man read the book about a dog. b. The dog ran away. |

L2 garden-path sentences are constructed in the same way as the sentence *while the man hunted the deer ran through the woods*, where the subject of the matrix clause (i.e., *the deer*) is the ambiguity region and the following verb (i.e., *ran*) is the reanalysis region. The pattern of non-garden-path sentences is the same as that of the garden-path sentences, except that the noun phrase in the ambiguity region cannot be attached to the proceeding verb due to semantic constraints, thereby causing no syntactic ambiguity. According to the rating results (1-highly implausible; 6-highly plausible) given by 20 Chinese-English bilinguals who didn't attend the formal experimental session, all sentences were regarded as plausible (garden-path: $M = 5.59$, $SD = 0.21$; non-garden-path: $M = 5.47$ $SD = 0.82$). Additionally, forty filler sentences were also presented in the task, including sentences with various types of syntactic structures.

The probe sentences of garden-path sentences included two types, half congruent with the globally correct interpretation and the other half congruent with the initial misinterpretation. For non-garden-path sentences, half of the probe sentences were consistent with the embedded clause of the experimental sentence and the other half were consistent with the whole sentence. Half of the fillers' probe sentences were congruent with the partial reading of the filler, while the rest were congruent with the global reading of the filler.

These sentences were classified into two lists based on the Latin Square design, with both lists including 30 garden-path sentences, 30 non-garden-path sentences, and 40 fillers. Half probe sentences were judged correct and the other half wrong. The assignment of congruent and incongruent responses to the left and right hands was counterbalanced across participants.

*Three Stroop tasks*

Inhibitory control can be measured by the tasks that demand the stopping of a prospective interfering response (Aron et al., 2014; Christ et al., 2007). Given that the predictive efficacy of the Stroop effects measured by these tasks may vary as shown by previous findings (Chen & Juola, 1982; Mägiste, 1984, 1985; Seidenberg, 1985; Wang et al., 2016; see the Discussion section for detailed reviews), the present study utilised L1 colour-word Stroop task, L2 colour-word Stroop task, and Number Stroop task to refine the assessment of inhibitory control.

*Colour-word Stroop task*

As shown in Table S1 (Supplementary Materials), L1 and L2 colour-word Stroop tasks share the same experimental design but differ in languages. In particular, this kind of task comprises two conditions—the congruent condition and the incongruent condition. In the former, the colour of the word matches its literal meaning, while it is the opposite in the latter. In this task, participants are requested to tell the ink colour of the word without focusing on its literal meaning. For example, if the word *red* is printed in blue ink, participants must name it "blue" instead of "red". Participants generally spend longer time in the incongruent condition than in the congruent one, in that they are distracted by the meaning of the word, which is more automatic processing than colour naming and, therefore, harder to inhibit (Macleod & MacDonald, 2000; Mitchell, 2005). This is called the Stroop effect. Statistically, it is calculated as the reaction-time difference between the incongruent and the congruent trials. One with a lower Stroop effect does better suppressing the automatically activated word meaning, indicating that he has better inhibitory control.

L1 Stroop task included congruent, incongruent, and neutral conditions (i.e., the word was not related to colour), each consisting of 50 trials. L2 Stroop task was identical to the L1 Stroop task, except that its materials were in English.

*Number Stroop Task*

Besides the colour-word Stroop task, there is also a non-verbal Number Stroop which features non-linguistic information (i.e., Roman numerals) of materials and experiment assignment (i.e., participants need to respond to either the sizes of two numbers instead of their numerical values or the numerical values regardless of their sizes) (e.g., Fitousi & Algom, 2006, 2018, 2020; Henik & Tzelgov, 1982). This task could reflect domain-general inhibitory control.

Number Stroop task included congruent, incongruent and two neutral conditions (i.e., the numeric values or the shapes were identical), with 30 trials in each condition, as shown in Table S1 (Supplementary Materials). The Arabic digits 2-8 were used as stimuli, each digit paired with a different digit that was smaller or bigger by two or with a numerically identical digit in a different physical size.

*English proficiency test*

The Oxford Quick Placement Test (paper and pen version) was adopted to measure participant's global English proficiency. It is a short diagnostic test in multiple-choice format, assessing reading, vocabulary and grammar skills. The total score of this test is 60. A score above 48 means the advanced level, and a score ranging from 30 to 47 represents the intermediate level. The English proficiency of our participants was measured by summing up the Oxford Quick Placement Test (60 questions, one score for each question). Higher scores indicate better English proficiency. The results showed that our participants were approximately at intermediate to advanced proficiency level ($M = 40.92$; $SD = 6.12$; range = 28-56).

### 2.3 Procedure

Participants completed the tasks in the order: Stroop-Reading-Stroop-Reading-Stroop, as shown in Figure S1. L1 Stroop task, L2 Stroop task and Number Stroop task were counterbalanced in the task order, so were the self-paced L1 and L2 garden-path sentence reading tasks. Sufficient break time between tasks was guaranteed. We kept the Stroop and Reading tasks apart in order to reduce practice or familiarity effects as much as possible.

### English proficiency test

The administration of the whole test took about 30 minutes. At the beginning of the experiment, participants were asked to complete the English Proficiency Test first. They needed to complete the test individually, seriously and honestly. Plagiarism was not allowed.

### Self-paced reading tasks

These tasks were performed via E-Prime 2.0. The sentences were randomly presented.

Prior to the experimental session, five practice trials were presented for participants to get familiar with the procedure. For each trial, following a fixation of "+" for 500 ms, the sentence segments, as manifested in Tables 1 & 2, were presented in a self-paced reading manner. Participants were asked to press the spacebar to read the sentence and respond to the probe sentence after the whole sentence was presented, by pressing one of the two keys on the keyboard (J/F) within 3000 ms to indicate whether the probe sentence was semantically congruent with the preceding sentence. The setting of the two keys for the congruent and incongruent probe sentences was counterbalanced among the participants. If no response was detected within this time limit, the next trial would begin. Participants were instructed to read carefully at a normal speed and respond to the probe sentences as quickly and accurately as possible. The RTs of the critical region and the ACCs of probe sentences were recorded. Participants were allowed to take a break in the middle of each task.

*Stroop tasks*

The Stroop tasks were performed via E-Prime 2.0. L1 Stroop task and L2 Stroop included 150 trials; Number Stroop task included 240 trials. Each task, where trials were randomly presented, included practice and experimental blocks. In the L1 and L2 Stroop tasks, the fixation of "+" was presented for 500 ms after the instruction, followed by a stimulus to which participants needed to respond within 3000 ms by pressing one of the two keys (J/F) to indicate whether the colour of printed words was red or blue. The assignment of these two keys to congruent and incongruent conditions were counterbalanced among participants. Participants were instructed to stay focused and respond to the colour as quickly and accurately as possible. The RTs and ACCs were recorded. If no response was detected within the time limit, the next trial would begin. Similarly, Number Stroop task contained two subtasks. Firstly, participants needed to compare the physical size of two numbers, ignoring their numerical value; if their size was the same, participants made judgment based on numerical value. Secondly, they needed to compare the numerical value, ignoring their physical size; if the value was the same, they made a judgment based on size. The sequence of both subtasks was counterbalanced. Participants were allowed to take a break after completion of each task.

## 2.4 Data analysis

Three participants' responses were removed from data analysis due to their low accuracy in the Stroop tasks (below 75%), similar to the benchmark adopted by Wang et al. (2016), leaving 39 participants in the final data sheet.

To determine the Stroop effects, we first cleaned the RTs by excluding incorrect responses, then removed the extreme RTs lower than 200 ms and higher than 2000 ms (Miller, 2023), and finally discarded RTs exceeding ±3SDs of each participant. The removing rate was 4.33% for L1 Stroop task, 3.82% for L2 Stroop task, as well as 4.96% for Number Stroop task. The Stroop effect for each Stroop task was calculated by subtracting the average RTs in the congruent condition from the average RTs in the incongruent condition. The larger the divergence, the bigger the Stroop effect, and the lower the inhibitory control ability. The original Stroop effects were then transformed to z-scores. They were treated as continuous variables in the original linear mixed-effect model; however, in the simple-effect analysis if significant interactions were found, these effects were divided into two-level groups by the median score of each factor, as detailed below. Each factor was coded via the dummy coding method (e.g., High L1 Stroop Effect = 1/2, Low L1 Stroop Effect = -1/2).

In L1 and L2 garden-path sentence reading task, longer RTs in the reanalysis region indicate ambiguity reanalysis, and ACCs measure lingering misinterpretation.

For the RT analysis, inaccurate responses were removed first. Next, extreme values with RTs below 200 ms and above 2000 ms were discarded. Finally, RTs exceeding ±3SDs of each participant were discarded. The removing rate was 11.41% for L1, and 16.37% for L2. Prior to our statistical analysis, the RTs were log-transformed (based on e) and Sentence Type was coded via the dummy coding method (i.e., garden-path = 1/2, non-garden-path = -1/2).

We conducted separate analyses of the L1 and L2 data, as the sentence materials differed due to the unique garden-path constructions of each language. Linear mixed-effects modellings of the RTs in the L1/L2 self-paced sentence reading tasks were conducted using R (R Core Team, 2021), with Sentence Type (garden-path, non-garden-path), L1 Stroop Effect and its interaction with Sentence Type, L2 Stroop Effect and its interaction with Sentence Type, and Number Stroop Effect and its interaction with Sentence Type as fixed effects, and participants and items as random effects. The treatment for the RT analyses of the L2 sentence reading task mirrored that of the L1 sentence reading task, with the exception that English Proficiency and its interactions were included as fixed effects in L2 models. In the simple-effect analysis, English Proficiency was divided into two levels based on the median score (High Proficiency vs. Low Proficiency) and then coded via the dummy coding method (i.e., High Proficiency =1/2, Low Proficiency = -1/2). Besides, the length of the critical words served as a covariate in the L2 model (Please see the model formula under Table 3 for details). This was the best-fit model for our data after model comparisons according to the AIC value (Akaike Information Criterion; the smaller, the better).

Besides, generalized mixed-effects modellings of the ACCs in the L1/L2 self-paced sentence reading tasks were conducted using R, with Sentence Type, L1 Stroop Effect and its interaction with Sentence Type, L2 Stroop Effect and its interaction with Sentence Type, and Number Stroop Effect and its interaction with Sentence Type as fixed effects, and participants and items as random effects. In the L2 model, Proficiency and its interaction with other factors were included as fixed effects (Please see the model formula under Table 4 for details). This was the best-fit model for our data after model comparisons according to the AIC value.

## 3. Results

### *3.1 Three Stroop tasks*

As shown in Table S2 (Supplementary Materials), in each Stroop task, the average RTs in the incongruent condition were significantly longer than the congruent condition. The difference in RTs between these two conditions is an indicator of the Stroop effect. The larger the Stroop effect, the smaller the inhibitory control. The Stroop effects obtained by the three Stroop tasks had no significant correlation, suggesting that they may measure different dimensions of inhibitory control (based on Pearson correlation criterion, please see Table S3 (Supplementary Materials) for details).

### *3.2 L1 self-paced sentence reading task*

*RT*

Mean RTs for the reanalysis region of Garden-path and Non-Garden-Path conditions are presented in Figure 1. Table 3 displays the results of the linear mixed-effects model of RTs. There was a significant main effect of Sentence Type with longer RTs for garden-path sentences (*M* = 431.64 ms, *SD* = 173.70 ms) than for non-garden-path sentences (*M* = 422.54 ms, *SD* = 174.30 ms). The effect of the Number Stroop was significant, suggesting longer reaction time for both garden-path and non-garden-path sentences with higher inhibitory control. No significant interaction was observed between Sentence Type and any Stroop effect.

**Table 3.** *Model parameters for the best-fitting model in the L1 & L2 sentence tasks (RT).*

| *Fixed effects* | L1 | | L2 | |
|---|---|---|---|---|
| | *F* | *p* | *F* | *p* |
| Type | **6.43** | **.011** | **19.05** | **<.001** |
| Length | - | - | **17.92** | **<.001** |
| L1 Stroop effect | 0.09 | .766 | 1.39 | .247 |
| L2 Stroop effect | 0.285 | .597 | 0.18 | .672 |
| Number Stroop effect | **7.01** | **.012** | 0.23 | .635 |
| Proficiency | - | - | 2.57 | .119 |
| Type×L1 Stroop effect | 1.59 | .207 | 0.46 | .497 |
| Type ×L2 Stroop effect | 0.37 | .542 | 0.28 | .596 |
| Type×Number Stroop effect | 1.48 | .224 | 0.79 | .375 |
| Type×Proficiency | - | - | 0.06 | .812 |
| Type×L1 Stroop effect×Proficiency | - | - | 0.29 | .747 |
| Type×L2 Stroop effect×Proficiency | - | - | 0.41 | .668 |
| Type×Number Stroop effect×Proficiency | - | - | 1.66 | .199 |

*Notes:* Model formula for LogRT in L1: LogRT ~ 1 + L1 Stroop effect + L2 Stroop effect + Number Stroop effect + Type + Type : L1 Stroop effect + Type : L2 Stroop effect + Type : Number Stroop effect +( 1 | Subject )+ ( 1 | Item ); Model formula for LogRT in L2: LogRT ~ 1 + L1 Stroop effect + L2 Stroop effect + Number Stroop effect +Proficiency+ Length+ Type + Type : L1 Stroop effect + Type : L2 Stroop effect + Type : Number Stroop effect+Type : Proficiency+Type : Proficiency:L1 Stroop effect+Type : Proficiency : L2 Stroop effect+Type : Proficiency : Number Stroop effect+( 1 | Subject )+ ( 1 | Item )
Significant results were marked in bold.

*ACC*

 Mean ACC rate of probe sentence comprehension of the two conditions are presented in Figure 1. Additionally, we calculated the percentages of errors stemming from two sources: initial misinterpretation (where the probe sentence aligns with the misinterpretation and participants made an affirmative judgment) and global misunderstanding (where the probe sentence aligns with the globally correct interpretation but participants made a negative judgment). The results, illustrated in Figure 2, indicate that initial misinterpretations accounted for the majority of errors (83.33%), suggesting that the accuracy rates predominantly reflect the extent of lingering misinterpretations. Table 4 presents the results of the generalized linear mixed-effects model of ACCs. A main effect of Sentence Type was significant with lower ACCs for garden-path sentences (*M* = 89.08%, *SD* = 0.31) than for non-garden-path sentences ($M = 97.21\%$, $SD = 0.16$). Moreover, the interaction of Sentence Type and L1 Stroop effect was significant. Simple-effect tests showed that in L1 Stroop effect$_{high}$ group (IC$_{low}$ group), ACCs were significantly lower for garden-path sentence ($M = 89.32\%$, $SD = 0.31$) than for non-garden-path sentences ($M = 97.85\%$, $SD = 0.15$; $\beta = 1.96$, $SE = 0.35$, $t = 5.68$, $p < .001$). In L1 Stroop effect$_{low}$ group (IC$_{high}$ group), ACCs were also significantly lower for garden-path sentences ($M = 89.70\%$, $SD = 0.30$) than non-garden-path sentences ($M = 96.59\%$, $SD = 0.18$; $\beta = 1.38$, $SE = 0.29$, $t = 4.75$, $p < .001$). These findings indicated that inhibitory control may not play a role in modulating the degree of lingering misinterpretation in L1.

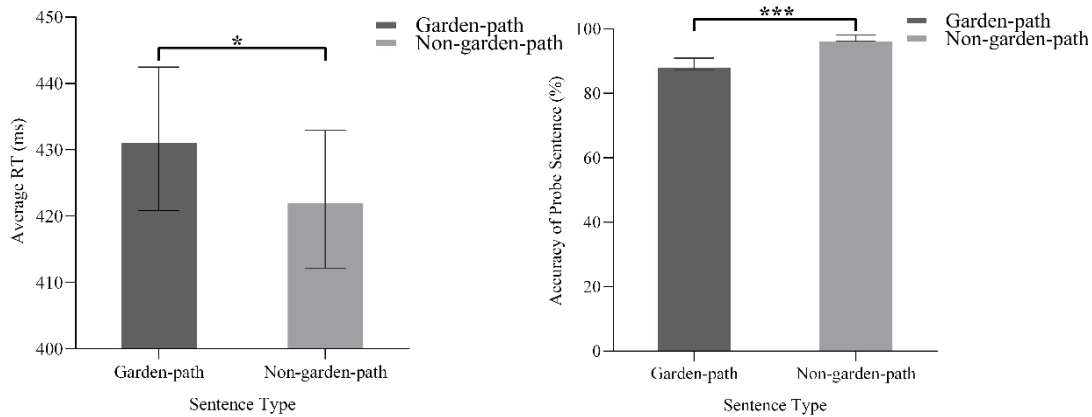

**Figure 1.** Mean RTs for the reanalysis region of garden-path and non-garden-path conditions (left panel); mean accuracies of probe sentences for garden-path and non-garden-path sentence conditions in the L1 self-paced sentence reading task (right panel).

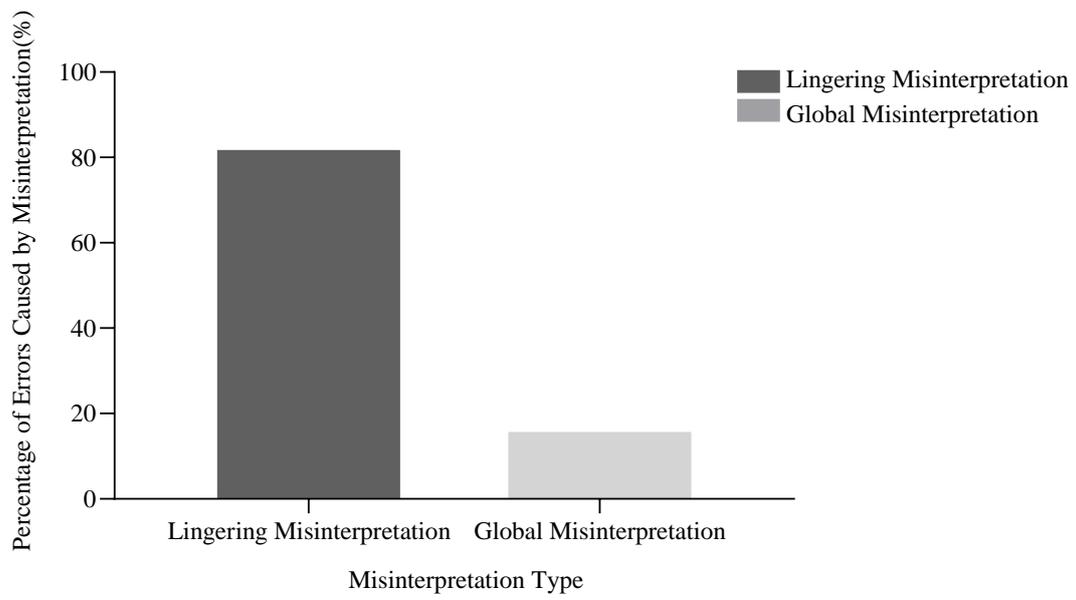

**Figure 2.** The percentage of total probe comprehension errors in the L1 self-paced sentence reading task attributed to initial misinterpretations and global misinterpretations.

**Table 4.** *Model parameters for the best-fitting model in the L1 & L2 sentence tasks (ACC).*

|  | L1 | | L2 | |
|---|---|---|---|---|
| *Fixed effects* | *F* | *p* | *F* | *p* |
| Type | **57.28** | **<.001** | **22.73** | **<.001** |
| Length | - | - | 0.57 | .451 |
| L1 Stroop effect | 2.98 | .084 | 0.28 | .596 |
| L2 Stroop effect | 0.19 | .666 | 0.01 | .915 |
| Number Stroop effect | 0.04 | .842 | 0.02 | .902 |
| Proficiency | - | - | 2.33 | .127 |
| Type×L1 Stroop effect | **5.84** | **.016** | 1.48 | .224 |
| Type ×L2 Stroop effect | 0.31 | .579 | 0.05 | .831 |
| Type×Number Stroop effect | 0.07 | .789 | 0.69 | .405 |
| Type×Proficiency | - | - | **6.22** | **.013** |
| Type×L1 Stroop effect×Proficiency | - | - | **6.17** | **.046** |
| Type×L2 Stroop effect×Proficiency | - | - | 0.11 | .946 |
| Type×Number Stroop effect×Proficiency | - | - | 0.02 | .989 |

*Notes:* Model formula for ACC in L1: ACC ~ 1 + L1 Stroop effect + L2 Stroop effect + Number Stroop effect + Type + Type : L1 Stroop effect + Type : L2 Stroop effect + Type : Number Stroop effect +( 1 | Participant )+ ( 1 | Item ); Model formula for ACC in L2: ACC ~ 1 + L1 Stroop effect + L2 Stroop effect + Number Stroop effect +Proficiency+ Length+ Type + Type : L1 Stroop effect + Type : L2 Stroop effect + Type : Number Stroop effect+Type : Proficiency+Type : Proficiency:L1 Stroop effect+Type : Proficiency : L2 Stroop effect+Type : Proficiency : Number Stroop effect+( 1 | Participant )+ ( 1 | Item )
Significant results were marked in bold.

### 3.3 L2 self-paced sentence reading task

*RT*

Figure 3 presents the mean RTs for the reanalysis region of garden-path and non-garden-path conditions. Table 3 shows the results of the linear mixed-effects model of RTs. There was a significant main effect of Sentence Type with longer RTs for garden-path sentences (*M* = 769.71 ms, *SD* = 373.38 ms) compared to non-garden-path sentences (*M* = 712.73 ms, *SD* = 339.10 ms). No significant interaction was observed between Sentence Type and any Stroop effects.

*ACC*

Figures 3 and 4 show the mean ACC rate for probe sentence comprehension of the two conditions, and the percentage of errors caused by initial misinterpretations and global misinterpretations. Initial misinterpretations accounted for most of the errors (i.e., 88.73%), suggesting that ACCs largely reflected the degree of misinterpretation. Table 4 displayed the results of the generalized linear mixed-effects model of ACCs. A significant main effect of Sentence Type with lower ACCs for garden-path sentences ($M = 87.86\%$, $SD = 0.33$) than for non-garden-path sentences ($M = 93.33\%$, $SD = 0.25$). The three-way interaction of Sentence Type, L1 Stroop Effect and Proficiency was significant. Simple-effect tests showed that Proficiency$_{low}$-IC$_{high}$ participants had significantly lower ACCs of garden-path sentences ($M = 82.50\%$, $SD = 0.38$) relative to non-garden-path sentences ($M = 92.22\%$, $SD = 0.27$; $\beta = 1.05$, $SE = 0.27$, $t = 3.84$, $p < .001$). Proficiency$_{high}$-IC$_{low}$ Participants had significantly higher ACCs of garden-path sentences ($M = 96.67\%$, $SD = 0.18$) relative to non-garden-path sentences ($M = 92.00\%$, $SD = 0.27$; $\beta = -1.02$, $SE = 0.43$, $t = -2.35$, $p = .019$). Yet, both Proficiency$_{low}$-IC$_{low}$ (garden-path: $M = 92.00\%$, $SD = 0.27$; non-garden-path: $M = 94.00\%$, $SD = 0.24$; $\beta = 0.15$, $SE = 0.54$, $t = 0.29$, $p = .775$) and Proficiency$_{high}$-IC$_{high}$ participants (garden-path: $M = 92.78\%$, $SD = 0.26$; non-garden-path: $M = 95.56\%$, $SD = 0.21$; $\beta = 0.50$, $SE = 0.52$, $t = 0.96$, $p = .338$) had no significant difference across conditions.

Table 5 shows the summary of the results of the simple-effect tests in the models of L1 and L2 sentence tasks.

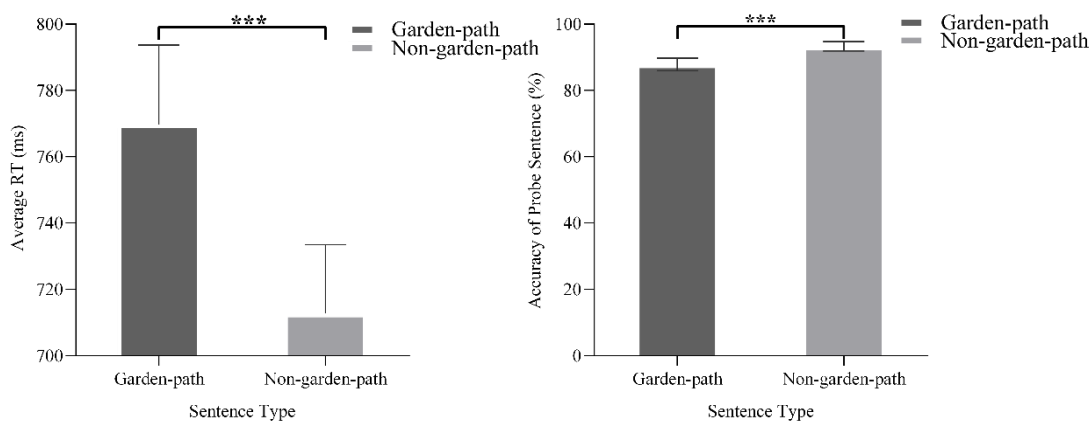

**Figure 3.** Mean RTs for the reanalysis region of garden-path and non-garden-path conditions (left panel); mean accuracies of probe sentences for garden-path and non-garden-path sentence

conditions in the L2 self-paced sentence reading task (right panel).

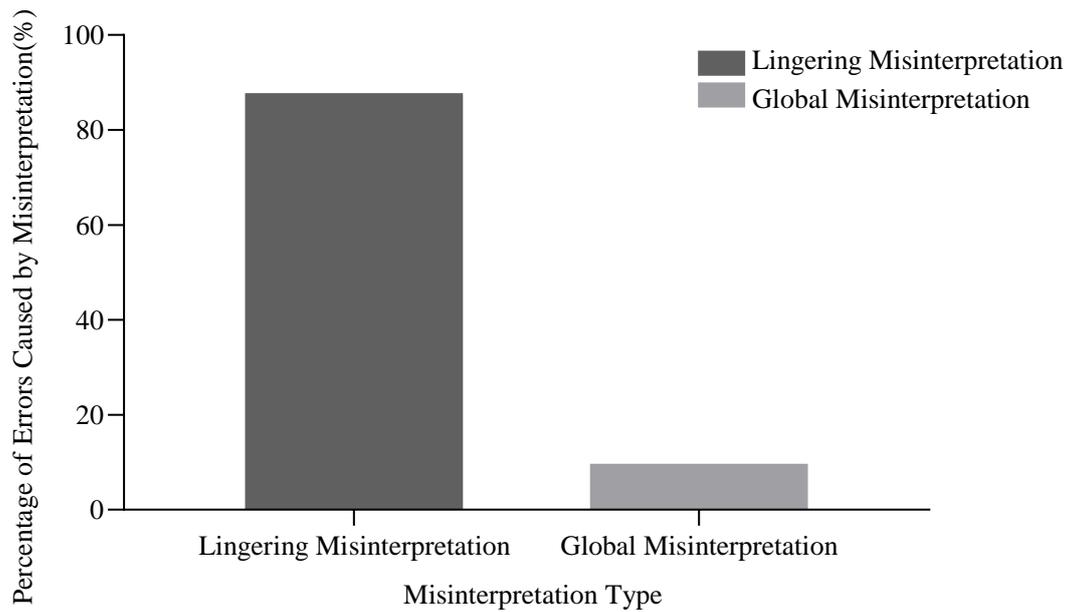

**Figure 4.** The percentage of total probe comprehension errors in the L2 self-paced sentence reading task attributed to initial misinterpretations and global misinterpretations.

**Table 5.** *Summary of the results of the simple-effect tests in the models of L1 and L2 sentence tasks.*

| | | L1 | | L2 | | | |
| | | | | *Proficiency$_{high}$* | | *Proficiency$_{low}$* | |
| *Simple effects* | | *RT* | *ACC* | *RT* | *ACC* | *RT* | *ACC* |
|---|---|---|---|---|---|---|---|
| Type-IC_L1 | Type-IC_L1$_{high}$ | NA | √ | NA | × | NA | √ |
| | Type-IC_L1$_{low}$ | NA | √ | NA | √ | NA | × |
| Type-IC_L2 | Type-IC_L2$_{high}$ | NA | NA | NA | NA | NA | NA |
| | Type-IC_L2$_{low}$ | NA | NA | NA | NA | NA | NA |
| Type-IC_Num | Type-IC_Num$_{high}$ | NA | NA | NA | NA | NA | NA |
| | Type-IC_Num$_{low}$ | NA | NA | NA | NA | NA | NA |

*Notes:* 1) IC = inhibitory control, IC_L1 = IC measured by L1 Stroop task, IC_L2 = IC measured by L2 Stroop task, IC_Num = IC measured by Number Stroop task. 2) ×means no significance of sentence type; √ means significance of sentence type found; NA means simple effect analysis was not conducted.

**4. Discussion**

This study employed a self-paced reading paradigm to investigate the role of inhibitory control in the processing of Chinese (L1) and English (L2) garden-path sentences in Chinese-English bilinguals. Our findings revealed a main effect of Sentence Type (garden-path vs. non-garden-path) in RT analyses for both languages, suggesting that processing garden-path sentences demands greater cognitive exertion, as reflected in longer reading times for the reanalysis region. These findings echo previous research (e.g., Ferreira & Henderson, 1991; Juffs & Harrington, 1996; Roberts & Felser, 2011; Sturt et al., 1999). Besides, lingering misinterpretations were observed in our study, as lower accuracies of probe sentence comprehension existed in both Chinese (L1) and English (L2) garden-path conditions relative to non-garden-path conditions, and most fof the inaccurate responses were attributed to lingering misinterpretations, consistent with findings in related studies (see also Christianson et al., 2001, 2006; Fujita & Cunnings, 2020; Jacob & Felser, 2016; Roberts & Felser, 2011). Critically, our findings contribute to the understanding of how Chinese garden-path sentences are processed, suggesting that the reanalysis and lingering misinterpretations in garden-path recovery are cross-linguistic phenomena.

Our key findings related to the role of inhibitory control are as follows: (1) Inhibitory control does not affect the processing of Chinese (L1) garden-path sentences; (2) The role of inhibitory control is modulated by L2 proficiency in the processing of English (L2) garden-path sentences. We will discuss our findings in detail in the following sections.

*4.1 None modulatory effect of inhibitory control on Chinese (L1) garden-path recovery*

To our knowledge, this study is the first to examine the role of inhibitory control in Chinese (L1) garden-path sentences. Our results showed that inhibitory control didn't modulate ambiguity reanalysis and lingering misinterpretations of Chinese (L1) garden-path sentences. Even though there was a significant interaction between Sentence Type and L1 Stroop effect, lingering misinterpretations of Chinese (L1) garden-path sentences were observed in both readers with high and low inhibitory control, as reflected in the lower accuracies of Chinese (L1) garden-path sentences compared with non-garden-path sentences. In other words, inhibitory control minimally influences the lingering misinterpretations in Chinese (L1). Since much of the evidence for the role of inhibitory control is derived from English contexts, the observed inconsistencies may be attributed to differences in ambiguity structures between Chinese and English garden-path sentences. While English garden-path sentences typically involve object/subject ambiguity, Chinese garden-path sentences often introduce the particle *de*, which necessitates reanalysis and clarifies the modification relationship between preceding and following phrases. As mentioned previously, the effect size of inhibition may be concerned with the strength of ambiguity present in sentences (Engelhardt & Ferreira, 2010). It is likely that the structure of Chinese garden-path sentences exhibits weak ambiguity and leads to little competition between interpretations. Consequently, parsers can easily recover from such sentences with limited reliance on inhibitory control. Another possibility is that the different characteristics of sentence processing between English and Chinese contribute to inconsistent findings. Specifically, Chinese tends to prioritize semantics over strict syntax, allowing meaning to be inferred from context, whereas English emphasizes syntactic clarity (Ming, 2023). This distinction may affect the cognitive resource allocation of inhibitory control while resolving syntactic ambiguity in different linguistic contexts. Of course, more investigation is needed to verify this finding.

### 4.2 The modulation of language proficiency and inhibitory control on the lingering misinterpretations of English(L2) garden-path sentences

Our results showed that inhibitory control modulated the processing of English (L2) garden-path sentences, aligning with previous research on native English speakers (Hsu & Novick, 2016; May & Scofield, 2024; Trueswell et al., 1999; Vuong & Martin, 2014). Notably, our findings suggest that garden-path recovery is a complex mechanism influenced by both language proficiency and inhibitory control. This supports the Model of Cognitive Control (Ness et al., 2023) and extends its applicability to the L2 context, where language proficiency has a substantial impact. We now discuss this three-way interaction among Sentence Type, L1 Stroop Effect and L2 Proficiency in detail. To be specific, first, L2 Proficiency$_{low}$-IC$_{high}$ participants had significantly stronger lingering misinterpretations of garden-path sentences compared to non-garden-path sentences. However, L2 Proficiency$_{low}$-IC$_{low}$ participants had comparable performances across garden-path and non-garden-path conditions. Second, while L2 Proficiency$_{high}$-IC$_{high}$ participants had little lingering misinterpretation shown by no significant difference in accuracies across conditions, L2 Proficiency$_{high}$-IC$_{low}$ participants had significantly higher accuracies in garden-path condition than non-garden-path condition. The accuracy differences between the garden-path condition and non-garden-path condition can be ranked as follows (Figure 5; for a summary of accuracy across conditions, refer to Table S4).

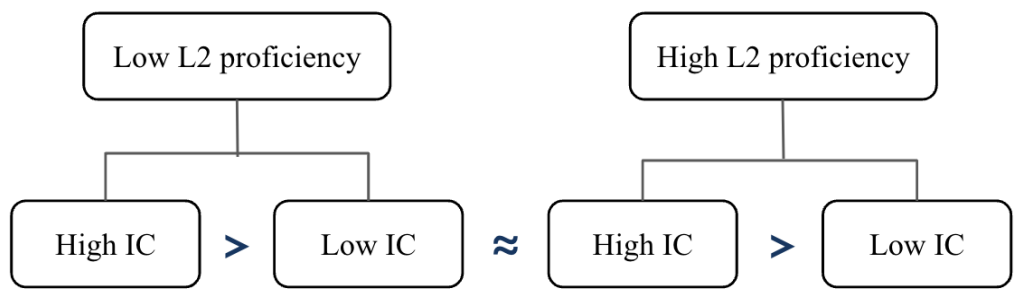

**Figure 5.** The ranking of accuracy differences between the garden-path condition and non-garden-path condition in L2 garden-path sentences.

These findings, at first glance, seem counter-intuitive. Yet, upon closer inspection, these results may be reasonable for the following reasons. Firstly, from the perspective of activation, for participants with low L2 proficiency, their representation of the globally correct interpretation may be complete but weak due to low L2 proficiency; instead, the initial misinterpretation, though abandoned, left a strong memory trace, which was unerased and prevailed during memory retrieval. Meanwhile, during memory retrieval, as inhibitory control refers to the ability to inhibit the weak competitor whose activation is relatively weak (Anderson, 2005), which is regarded "wrong" interpretation in order to reach the "right" interpretation, we conjecture that participants with low L2 proficiency but high inhibitory control may sometimes inhibit the assumed "wrong" but globally correct representation, which has a weaker activation, while retrieving the regarded "correct" but misinterpretation from their memory trace, leading to a higher rate of lingering misinterpretations (i.e., "more wrong") as manifested by significantly lower ACC in the garden-path condition than the control condition. In contrast, for those with low L2 proficiency and low inhibitory control, they exhibited a lesser degree of lingering misinterpretations, as evidenced by the averaged ACC in the garden-path condition being lower than the control condition, though the difference was not statistically significant. We speculate that this lesser degree of lingering misinterpretations may result from less inhibition of the assumed "wrong" but globally correct representation, which has weaker activation. These findings support and provide new insights into the Memory Encoding and Retrieval Interference theory (Cunnings, 2017). As Cunnings argues, L2 learners construct fully specified syntactic structures but are more susceptible to retrieval interference from competing memory traces than L1 speakers. Consequently, they may rely more heavily on cognitive control during memory retrieval, where successful comprehension depends on accessing relevant information. Our findings are aligned with this account as we found that L2 learners experience lingering misinterpretations, and that inhibitory control modulates these effects.

Secondly, L2 parsers with high L2 proficiency might facilitate a strong and complete representation of the globally correct interpretation, and abandon the initial misinterpretation. L2 Proficiency$_{high}$-IC$_{low}$ participants exhibited no lingering misinterpretations. Notably, the ACC in the garden-path condition was significantly higher than that in non-garden-path condition. This suggests that the diminished lingering misinterpretations may result from the representation of the globally correct interpretation being strong enough to override the interference from distracting memory traces. In the meantime, non-garden-path sentences were still processed at a normal way, resulting in higher accuracy in memory retrieval in the garden-path condition instead. However, L2 proficiency$_{high}$-IC$_{high}$ showed neither greater ACC in the garden-path condition nor significant lingering misinterpretations, as indicated by comparable ACC between the garden-path and control conditions. This outcome likely arises when the activation of the globally correct interpretation in some trials (though not all) is only marginally stronger than that of the misinterpretation. Given their high inhibitory control, these participants may exert influence selectively on these trials, achieving an equilibrium of matched performance between the two conditions. Overall, even at high levels of L2 proficiency, the modulation of inhibitory control on lingering misinterpretations persists. These findings are broadly consistent with the role of inhibitory control observed in L1 English speakers (Hsu & Novick, 2016; May & Scofield, 2024; Trueswell et al., 1999; Vuong & Martin, 2014), though its role is subject to language proficiency in the L2.

Overall, these findings provide supporting evidence from a bilingual perspective for the *unerased false memory trace account* in the Good Enough Processing model (Christianson et al., 2001; Ferreira et al., 2001, 2002; Ferreira & Patson, 2007; Huang & Ferreira, 2021; Qian et al., 2018; Slattery et al., 2013; Sturt, 2007). In particular, given that there is no lingering misinterpretation among those with L2 Proficiency$_{low}$-IC$_{low}$ and L2 Proficiency$_{high}$-IC$_{high}$, our results show that participants can establish correct and complete representations after reanalysis, regardless of levels of L2 proficiency. As for the different degree of lingering misinterpretations among groups, it could be attributed to the relatively divergent activations of the established representation and the memory trace, and the interactive modulation of L2 proficiency and inhibitory control, as discussed above. Regarding the long-standing puzzle of why readers fail to fully erase the memory traces of initial misinterpretations, this study suggests that this issue is related to individual factors such as language competence, inhibitory control, etc. Future studies should investigate additional individual factors that may be involved and examine how these factors interact to influence garden-path recovery.

Based on the Model of Cognitive Control (Ness et al., 2023), we tend to suggest a dynamic view of the processing of garden-path sentences. From the beginning, the reader may hold all possible interpretations, with misinterpretation supported by the most reliable evidence (i.e., constraints) in a more active state. Upon reading the reanalysis word, he may revise the initial constraints and raise the activity level of the globally correct interpretation with the help of his syntactic and semantic knowledge whose levels are mainly represented by his language proficiency. The conflict between interpretations is resolved by competition. During this process, linguistic knowledge guides the relative activation of those interpretations, while inhibitory control functions as a tool to inhibit the interpretation that is supported by weaker evidence. By the end of the processing stage, two interpretations—each at different activation levels—are stored in memory. Lingering misinterpretations, as explained by the *unerased false memory trace account*, arise from the strong memory trace of the initial misinterpretation. During memory retrieval, inhibitory control targets the weaker competitor, whether it is the misinterpretation or the globally correct interpretation, to suppress its activation. Of course, this point of view needs to be examined and provided more empirical evidence in future research.

Besides, we didn't observe the modulatory effect of inhibitory control and L2 proficiency on L2 garden-path reanalysis (as shown by the RT analysis), which aligns with recent work by Xie et al. (2022). This was contrary to our original hypothesis, suggesting that the relationship between these factors is more nuanced than previously assumed. The modulation was instead observed in the subsequent retrieval phase, indicating that inhibitory control and L2 proficiency interact to suppress the intruding memory trace during the later stage of sentence processing. However, it cannot exclude the possibility that the interaction of the two factors may modulate the early reanalysis process, which would require techniques with higher temporal resolution, such as eye-tracking, to gain deeper insights.

### *4.3 A comparison of the three Stroop tasks*

A notable aspect of this study is the adoption of three types of Stroop tasks to identify which most effectively reveals the role of inhibitory control in garden-path comprehension. Our results showed that the linguistic domain Stroop task (i.e., L1 Stroop task) may better measure inhibitory control in language processing studies than domain-general Stroop task (i.e., Number Stroop task). This echoes the results of Vuong and Martin (2014), which showed that the time to revise garden-path interpretations was linked to resolving verbal Stroop interference, but not non-verbal Stroop interference. Collectively, these results suggest a potentially domain-specific role of inhibitory control in revising misinterpretations during sentence comprehension.

Among the linguistic domain Stroop tasks, our findings indicate that L1 Stroop task is the most effective in predicting inhibitory control during garden-path recovery, while the L2 Stroop task exhibits reduced efficacy. Existing literature suggests that the intensity of the Stroop effect can vary significantly across tasks due to the specific linguistic information involved. The orthographic variation hypothesis suggests that Chinese orthography would generate a more significant Stroop effect than English counterparts (e.g., Chen & Juola, 1982; Seidenberg, 1985). Also, the impact of language proficiency on measuring the Stroop effect remains noteworthy. Mägiste (1984, 1985) found that their bilingual participants initially showed more interference when responding in L2, but with their L2 proficiency enhancing, interference caused by L1 and L2 reached a point of equivalence. Conversely, Wang et al. (2016) demonstrated that a comparable Stroop effect between proficient and non-proficient bilinguals was observed in L2 Stroop task (English). Yet, in L1 Stroop task (Chinese), proficient Chinese-English bilinguals presented a smaller Stroop effect than non-proficient bilinguals. Despite this, previous studies typically measured inhibitory control through a single task, without considering the subtle differences among the three Stroop tasks. Limited research has delved into which Stroop task performs best in reflecting one's inhibitory control to a large extent under certain circumstances (e.g., L1 processing or

L2 processing).

Our findings contribute new insights to this area. We demonstrate that the L1 Stroop task more effectively represents the inhibitory control involved in sentence processing, whereas the L2 Stroop task is less effective. This aligns with Mägiste's assertion that L2 proficiency can interfere with the Stroop effect, indicating that linguistic factors—such as second language proficiency and the competitive dynamics between L1 and L2—may influence the measurement of inhibitory control.

To sum up, these three Stroop tasks seemed to measure different dimensions of inhibitory control. The L1 and L2 Stroop tasks primarily assess inhibitory control within the linguistic domain and L1 Stroop task performs best in predicting inhibitory control recruited for garden-path recovery, whereas the Number Stroop task likely reflects domain-general inhibitory control, which may not significantly contribute to resolving temporary ambiguities. Future research is necessary to verify these findings and establish the optimal contexts for employing each Stroop task in the assessment of inhibitory control.

## 5. Conclusion

In summary, this study pioneers the examination of inhibitory control in the context of Chinese (L1) garden-path sentences. Additionally, it provides critical insights into the interplay among inhibitory control, language proficiency, and the processing of garden-path sentences in English (L2), thereby supporting and extending the Model of Cognitive Control (Ness et al., 2023). Our findings indicate that inhibitory control does not significantly influence recovery from garden-path sentences in Chinese (L1), suggesting a potential divergence in how different languages resolve syntactic ambiguity. Specifically, Chinese's reliance on semantic context rather than strict syntactic rules may diminish the need for inhibitory control during ambiguity resolution.

In contrast, the results for English L2 learners affirm the complex interaction between language proficiency and inhibitory control. Participants with low L2 proficiency but high inhibitory control showed lingering misinterpretation. This suggests that their initial misinterpretations left a strong memory trace, and their high inhibitory control may suppress the weaker, albeit correct, interpretations. In contrast, those with high proficiency exhibited none, suggesting that their strong representation of globally correct interpretations overrides competing memory traces. This supports the Memory Encoding and Retrieval Interference theory, highlighting the crucial role of inhibitory control in managing competing memory traces during memory retrieval among L2 parsers. Moreover, our comparison of three types of Stroop tasks reveals that the L1 Stroop task is the most effective for measuring inhibitory control within linguistic contexts. This finding indicates the domain-specific nature of inhibitory control involved in garden-path sentence processing and underscores the importance of carefully selecting measurement tools in future research. Overall, these insights contribute to our understanding of the cognitive mechanisms underlying temporary ambiguities and highlight the necessity for further investigation into how individual differences in cognitive control and language proficiency affect sentence processing across different languages.

**Supplementary material:**

**Supplementary table 1.** *Materials of experimental stimulus in the three types of Stroop tasks.*

| Conditions | L1 Colour Stroop | L2 Colour Stroop | Number Stroop |
|---|---|---|---|
| Congruent | 红 蓝 | **red blue** | Value: $4$ $_2$ |
| | | | Size: $2$ $_4$ |
| Incongruent | 红 蓝 | **red blue** | Value: $2$ $_4$ |
| | | | Size: $4$ $_2$ |
| Neutral | 最 最 | **most most** | Value: $24$ $_{/\ 24}$ |
| | | | Size: $2$ $_2$ |

*Note*: "Value" means comparing the value in the task; "Size" means comparing the size in the task.

**Supplementary table 2.** *Mean reaction times and Mean Stroop effects in each condition and Stroop task.*

| Task | Type | Mean RT(SD) (ms) | Mean ACC(SD)(%) | Stroop effect |
|---|---|---|---|---|
| L1 Stroop task | Congruent | 462.88(143.34) | 97.85(0.15) | 36.73 |
| | Incongruent | 499.61(193.01) | | |
| L2 Stroop task | Congruent | 444.16(129.10) | 98.28(0.13) | 14.64 |
| | Incongruent | 458.80(145.05) | | |
| Number Stroop task | Congruent | 593.36(214.43) | 97.65(0.15) | 67.22 |
| | Incongruent | 660.58(236.56) | | |

**Supplementary table 3.** *Pearson correlation coefficient of three kinds of Stroop effect.*

| | L1 Stroop Effect | L2 Stroop Effect | Number Stroop Effect |
|---|---|---|---|
| L1 Stroop effect | 1 | | |
| L2 Stroop effect | .17 (.29) | 1 | |
| Number Stroop effect | .03 (.86) | -.02(.90) | 1 |

*Note*: None of the coefficients was significant. Please see the $p$ value in the parentheses.

**Supplementary table 4.** *Summary of accuracy in the garden-path condition and non-garden-path condition in L2 garden-path sentences.*

|  | Garden-path (%) | Non-garden-path (%) |
|---|---|---|
| Low L2 proficiency + High IC | $M = 82.50$, $SD = 0.38$ | $M = 92.22$, $SD = 0.27$ |
| Low L2 proficiency + Low IC | $M = 92.00$, $SD = 0.27$ | $M = 94.00$, $SD = 0.24$ |
| High L2 proficiency + High IC | $M = 92.78$, $SD = 0.26$ | $M = 95.56$, $SD = 0.21$ |
| High L2 proficiency + Low IC | $M = 96.67$, $SD = 0.18$ | $M = 92.00$, $SD = 0.27$ |

**Data availability：** All materials, data and analysis code of the current study are available via the Open Science Framework (https://osf.io/uckfg/). None of this study's design and analysis were pre-registered.

**Competing interests:** The authors declare none.